\documentclass[runningheads]{llncs}
\usepackage{graphicx}
\usepackage{subcaption}
\usepackage{tabularx}
\usepackage{amsmath}
\def\lc{\left\lceil} % LEFT CEILING 
\def\rc{\right\rceil} % RIGHT CEILING

\begin{document}
\title{Mixed-Initiative Level Design with RL Brush}
% \author{The Anonymous Sultan}
\author{Omar Delarosa\inst{1} \and 
Hang Dong\inst{1} \and 
Mindy Ruan\inst{1} \and 
\\Ahmed Khalifa\inst{1, 2} \and
Julian Togelius\inst{1,2}}

% \authorrunning{A. Sultan}
\authorrunning{O. Delarosa et al.}
% \institute{Anonymous School, Anonymous Capital, Anonymous Empire
% \email{anonymous@anonymousempire.com}}
\institute{New York University, Brooklyn NY 11201, USA
\email{\{omar.delarosa,hd1191,mr4739,ahmed.khalifa\}@nyu.edu}\and
modl.ai, Copenhagen, Denmark\\
\email{julian@togelius.com}}

\maketitle              % typeset the header of the contribution
\begin{abstract}
This paper introduces \textit{RL Brush}, a level-editing tool for tile-based games designed for mixed-initiative co-creation.  The tool uses reinforcement-learning-based models to augment manual human level-design through the addition of AI-generated suggestions. Here, we apply \textit{RL Brush} to designing levels for the classic puzzle game \textit{Sokoban}. We put the tool online and tested it in 39 different sessions. The results show that users using the AI suggestions stay around longer and their created levels on average are more playable and more complex than without.

\end{abstract}
\section{Introduction}
Modern games often rely on procedural content generation (PCG) to create large amounts of content autonomously or with limited human input. PCG methods can achieve many different design goals as well as enable particular aesthetics. Incorporation of PCG methods can streamline time-intensive tasks such as designing thousands of unique tree assets for a forest environment. By off-loading these tasks to AI, the time constraints put on game developers and content creators can be relaxed freeing them up to work on other tasks for which AI may be less-well suited.  Through such blending of AI and the human touch a system of human and AI co-creation yields not only unique game content the human designer may not have even considered alone but also enables new creative directions~\cite{shaker2016procedural}.

In Procedural Content Generation via Reinforcement Learning, or \textit{PCGRL} \cite{khalifa2020pcgrl}, levels are first randomly generated and then incrementally improved. The generated levels are initially good enough that they could--though unlikely good enough that they \textit{would}--be used by a human designer. That reluctance to use a level could arise from a level's misalignment with the human designer's needs and they would likely have to keep generating new levels until they find one that is satisfactory. Generally speaking, the human designer exerts minimal control over the resulting level's features and may end up generating many just to find one that suits their needs.

In order to make this level generation method more compatible with a human designer's workflow, we leverage the incremental nature of PCGRL in building a mixed-initiative level-editing tool. This paper presents \textit{RL Brush}, a human-AI collaborative tool that balances user-intent and AI model suggestions. \textit{RL Brush} allows a human designer to create levels as they please while continuously suggesting incremental modifications to improve the level using an ensemble of AI models. The human designer may choose to accept suggestions as they see fit or reject them.  The tool thereby aims to assist and empower human designers to create levels that are good, unique, and suitable to the user's objectives. 

% TODO: add a brief blurb here layout out what the topic/thesis of the paper is going to be.  i.e. we are presenting a tool

% TODO: Add a paragraph outlining the layout of the rest of the paper

\section{Related Work}
%Ahmed: Talk in general about PCG here and say near the end that you will focus on Reinforcement learning

Procedurally generated content has been used in games since the early 1980s. Early PCG-based games like \textit{Rogue} (Michael Toy, 1980) used PCG to expand the overall depth of the game by generating dungeons as well as coping with the hardware limitations of the day \cite{shaker2016procedural,yannakakis2018artificial}. This section will lay out more contemporary applications and methods of generating game content procedurally, with a focus on the use of reinforcement-learning-based approaches.

% TODO: mention GAN editor

% TODO: put more emphasis on the work of Alvarez, Evo Dungeon Designer. (see bib)

% TODO: edit the language so that it's mostly only making relevant definitions, omit stuff that was only relevant to our proposal.
\subsection{PCG via Reinforcement Learning}
Reinforcement Learning (RL) is a Machine Learning technique where, typically, an agent takes action in an environment at each time-step and desirable actions are reinforced, interpreted as state and reward, from the environment~\cite{sutton1998introduction}. Most of the RL work in games focuses on playing.  We suspect it may be due to the direct and easy way of representing the game playing problems as Markov decision processes. On the other hand, representing content generation as a Markov decision process poses challenges and hence could explain the disproportionally smaller number of works using RL in game content generation problems.

Of the existing works describing RL approaches to game content generation, a few approaches stand out and demonstrate the breadth of possibilities. Chen et al.~\cite{chen2018q} demonstrate using Q-learning to create a card deck for collectable cards games. Guzdail et al.~\cite{guzdial2019friend} have shown how active learning can be used to adjust an AI agent to adapt to user choices while creating levels for Super Mario Bros. 
% Ahmed: mention the other PCGRL stuff from my paper such as playing simcity by Sam Earle, Generating Deck of cards using Q-Learning, and Matthew Guzdial work of collaboration. Then enter to the PCGRL by saying that this paper defined that technique and go on as you are saying.
PCGRL\cite{khalifa2020pcgrl} introduces reinforcement learning into level generation by seeing the design process as a sequential task. Different types of games provide information on the design task as functions: an evaluation function that assesses the quality of the design and a function that determines whether the goal is reached. RL agents that \textit{play} out the content generation task defines the state space, action space, and transition function. For typical 2D grid based games, the state can be represented as a 2D array or 2D tensor. Agents of varying representation observe and edit the map using different patterns. The work demonstrates how three types of agents, namely \textit{narrow}, \textit{turtle} and \textit{wide}, can respectively edit tiles in a sequential manner, move on the map in a turtle-graphics-like way and modify the passed tiles, or have control to select and edit any tile in the entire map. %Ahmed: Maybe add a little explanations about these representations (narrow, wide, turtle) so people can understand what they are as I suspect they are explained anywhere else and they are mentioned in your method as three different models.

\subsection{PCGRL Agents}

The three RL-based level-design agents introduced in \textit{PCGRL} \cite{khalifa2020pcgrl} \cite{bhaumik2019tree} as \textit{narrow}, \textit{turtle} and \textit{wide} have origins in search-based approaches to level-generation, however the primary focus of the subsequent sections will be on their RL-based implementations.  This section describes these three canonical agent types.

\subsubsection{Narrow}
The \textit{narrow} agent observes the state of the game and a \textit{location} $(x,y)$ on the 2D-array grid representation of the game level.  Its action space consists of a \textit{tile-change} action: whether to make a change or not at location $(x,y)$ and what that change would be.

\subsubsection{Turtle}
Inspired by turtle graphics languages such as \textit{Logo} \cite{goldman2004turtle} \cite{khalifa2020pcgrl}, \textit{turtle} agent also observes the state of the grid as a 2D array and a \textit{location} $(x,y)$ on that grid.   Like \textit{narrow} agent, one part of its action-space is defined as a \textit{tile-change} action.  Unlike \textit{narrow}, its action space also includes a \textit{movement-action} in which the agent changes the agent's current position on the grid to $(x',y')$ by applying a 4-directional translation on its \textit{location} moving it either \textit{up}, \textit{down}, \textit{left} or \textit{right}.

\subsubsection{Wide}
The \textit{wide} agent also observes the state of the grid as a 2D array.  However, it does not take a \textit{location} parameter.  Instead, its action space selects a \textit{location} on the grid $(x,y)$ as the \textit{affected location} and a \textit{tile-change} action.

\subsection{PCG via Other Machine Learning Methods}

In addition to RL, other machine learning (ML) approaches have also been applied to procedural content generation and mostly based on supervised or unsupervised learning; the generic term for this is Procedural Content Generation via Machine Learning (PCGML)~\cite{summerville2018procedural}.  \textit{Mystical Tutor}  \cite{summerville2016mystical}, an iteration on the Twitter bot \texttt{@RoboRosewater}, generates never-before-seen \textit{Magic: The Gathering} cards using an Long short-term memory (LSTM) neural network architecture.  While Torrado et al. \cite{torrado2019bootstrapping} demonstrate that \textit{Legend of Zelda} (Nintendo, 1986) levels can be generated using generative adversarial networks (GAN). %Ahmed: Give some example of PCGML stuff. You can say "Researchers have experiment with a lot of different methods such GANs~\cite{couple of papers}, AutoEncoders~\cite{couple of papers}, N-Grams~\cite{papers}, Markov Chains~\cite{papers}, LSTMs~\cite{papers} etc."
However, one distinction that arises when comparing PCGRL with other types of PCGML: PCGRL does not strictly require ahead-of-time training data. RL-based models utilize a system of reward functions instead, which can be either manually designed or themselves learned. % Ahmed: you should mention that it needs on the other hand a reward function because you can always learn the reward function later.
PCGRL's system of incremental approach to level-generation also distinguishes it from more holistic ML approaches such as many GAN-based PCGML approaches. % Ahmed: you should know that Markov Chains, N-Grams, LSTMs, work are incremental modifications. I think you should soften the claim. Because these work are the inspiration for PCGRL in its incremental nature. GANs and AutoEncoders and other non incremental models are what you are talking about. Try to soften it 
At each time step, the agent takes an action such as moving to or selecting a certain position (for example, in 2D grid space) or changing the tile at the current position. This characteristic of PCGRL makes it well-suited for mixed-initiative design. %Ahmed: May be talk about Latent Vector Evolution by Vanessa Volz in evolving mario using GANs.

\subsection{PCG via Mixed-initiative Level Design}
In mixed-initiative design, the human and an AI system work together to produce the final content~\cite{yannakakis2014mixed,zhu2018explainable}. Multiple mixed-initiative tools for game content creation have been developed in recent years. \textit{Tanagra}\cite{smith2010tanagra} is a prototype mixed-initiative tool for platformer level design in which AI can either generate the entire level or fill in the gaps left by human designers. \textit{Sentient Sketchbook}\cite{liapis2013sentient} is a tool for designing a \textit{Starcraft}-like (Blizzard, 1998) strategy game. Users can sketch in low-resolution and create an abstraction of the map in terms of player bases, resources, passable and impassable tiles. It uses feasible-infeasible two population GA (FI-2pop GA) for novelty search and generates several map suggestions as users are sketching. An example of a mixed-initiative PCG tool that generates levels for a specific game is \textit{Ropossum}, which creates levels for the physics-based puzzle game \emph{Cut the Rope}, based on a combination of grammatical genetic programming and logic-constrained tree search
~\cite{shaker2013ropossum,shaker2013evolving}. Another such example is the mixed-initiative design tool for the game \emph{Refraction}, which teaches fractions; that tool is built around a constraint-solver which can create puzzles of specific difficulty~\cite{butler2013mixed}.

More recently, Alvarez et al.\cite{alvarez2019empowering} introduced Interactive Constrained MAP-Elites for dungeon design, which offers similar suggestion-based interaction supported by MAP-Elites algorithm and FI-2pop evolution. Guzdial et al.\cite{guzdial2018co} proposed a framework for co-creative level design with PCGML agents. This framework uses a level editor for \textit{Super Mario Bros} (Nintendo, 1985), which allows the user to draw with a palette of level components or sprites. After finishing one turn of drawing, the user clicks the button to allow the previously trained agent to make additions sprite-by-sprite. This tool is also useful for collecting training data and for evaluating PCGML models. In a similar vein, Machado et al. used a recommender system trained on databases of existing games to recommend game elements including sprites and rules across games
~\cite{machado2019pitako}.

\section{Methods}

\begin{figure}
    \centering
    \includegraphics[width=0.8\linewidth]{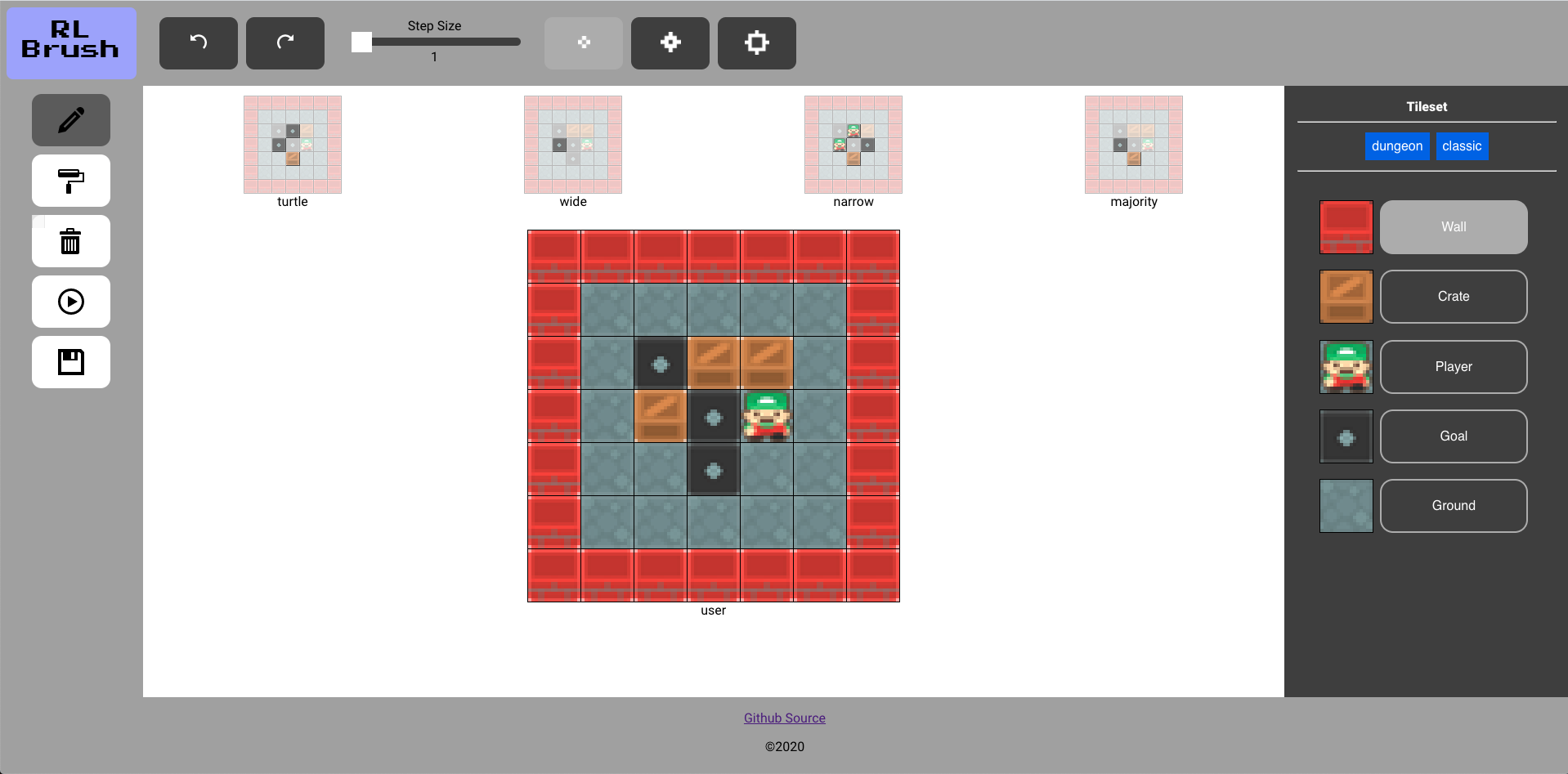}
    \caption{RL Brush screenshot of the Sokoban level editor.}
    \label{fig:rlbrush_editor}
\end{figure}

This section introduces \textit{RL Brush}, a mixed-initiative level-editing tool for tile-based games that uses an ensemble of trained level-design agents to offer level-editing suggestions to a human user. Figure~\ref{fig:rlbrush_editor} shows a screenshot of the tool~\footnote{https://rlbrush.app/}. The present version of \textit{RL Brush} is tailored for building levels for the classic puzzle game \textit{Sokoban} (Thinking Rabbit, 1982) and generating suggestions interactively.

% TODO: Describe sokoban here
\subsection{Sokoban}
\textit{Sokoban}, or ``warehouse keeper'' in Japanese, is a classic 2-D puzzle game in which the player's goal is to push boxes to their designated locations within an enclosed space (called goals). The player can only push boxes horizontally or vertically. The number of boxes is equal to the number of designated locations. The player wins when all boxes are in the correct locations.

\subsection{RL Brush}

In the spirit of human-AI co-creation of tools like \textit{Evolutionary Dungeon Designer} \cite{alvarez2018assessing} and \textit{Sentient Sketchbook} \cite{liapis2013sentient}, \textit{RL Brush} interactively presents suggested edits in to a human level creator, 4 suggestions at a time. Instead of using search-based approaches to generate the suggestions \textit{RL Brush} utilizes the reinforcement-learning-based level-design agents presented by \cite{khalifa2020pcgrl}.  \textit{RL Brush} builds on the work introduced by \textit{PCGRL} \cite{khalifa2020pcgrl} by combining user-interactions with the level-designing \textit{narrow}-, \textit{turtle}- and \textit{wide}-agents and an additional \textit{majority}, meta-agent into a human-in-the-loop\footnote{These are a subclass of AI-based systems that are designed around human interaction being one of their components}, interactive co-creation system.
%julian maybe we should describe these agents somewhere? -- DONE, in separate section

\subsection{Architecture Overview}

\begin{figure}
    \centering
    \includegraphics[width=0.8\linewidth]{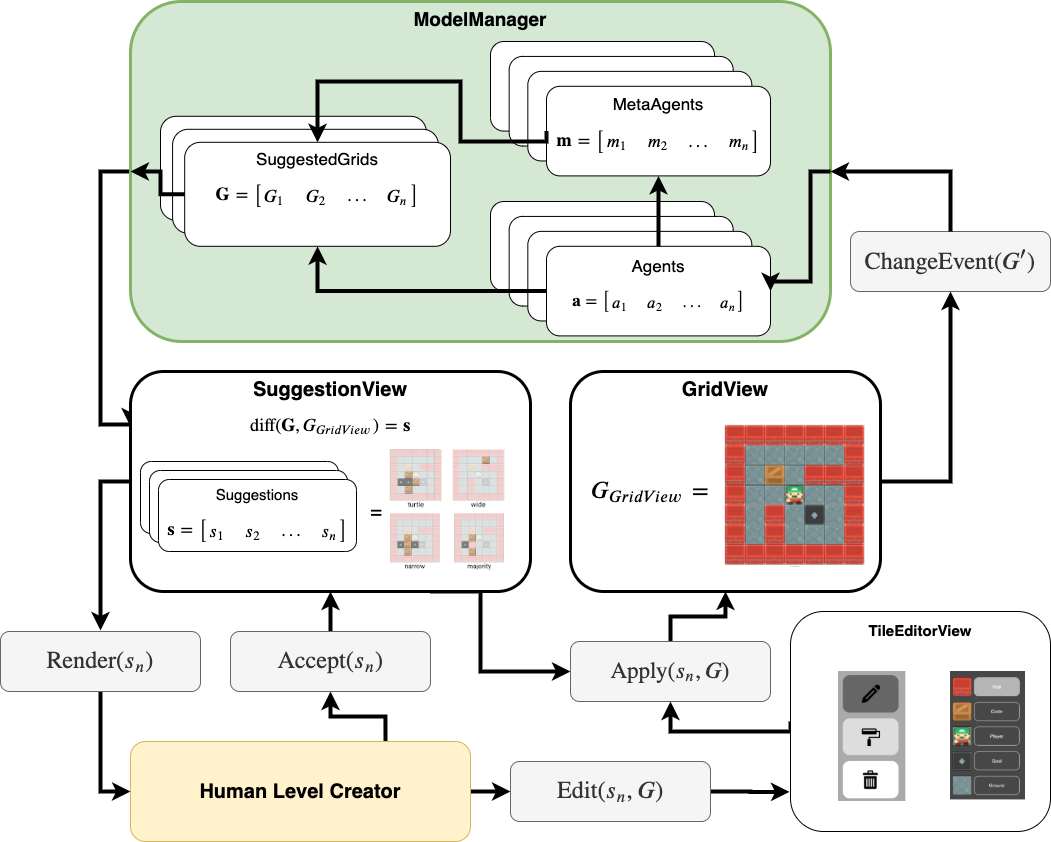}
    \caption{RL Brush System Arcihtecture.}
    \label{fig:architecture_diagram}
\end{figure}

%Ahmed: I changed the text to be more scientific than a software documentation. Please validate these changes
Figure~\ref{fig:architecture_diagram} shows the system architecture for our tool RL Brush. The system consists of 4 main components:
\begin{itemize}
    \item \textbf{GridView} is responsible for rendering and modifying the current level state.
    \item \textbf{TileEditorView} allows the user to select tools to edit the current level viewed in the GridView.
    \item \textbf{SuggestionView} shows the different AI suggestions from the current level in the GridView.
    \item \textbf{ModelManager} updates all the suggestions viewed in SuggestionView if the current level changed in the GridView.
\end{itemize}

The user can edit the current level ($G$) either by selecting a suggestion from the $SuggestionView$ or by using a tool from the $TileEditorView$ and modifying directly the map. This change emits a $ChangeEvent$ signal to the $ModelManager$ component with the new grid ($G'$). The ModelManager runs all the AI models and collects their results and send the results back to the $SuggestionView$. The $ModelManager$ will be described in more details in subsequent section.

% The co-creation process described in figure~\ref{fig:architecture_diagram} centers around a $\texttt{GridView}$ that renders a game level. This level is usually represented as a an integer grid $G$ where each integer represents a game entity. For example, figure~\ref{fig:tileset} shows the different game entities for our target game ``Sokoban''. The user can edit the $\texttt{GridView}$  by sending $\texttt{Apply}$ event objects to it either from the $\texttt{TileEditorView}$ (figure~\ref{fig:tileset}) or the $\texttt{SuggestionView}$ (figure~\ref{fig:suggestions_ui}). The $\texttt{GridView}$ apply these events and renders their resulting mutation to the grid and emitting a $\texttt{ChangeEvent}$ event to the $\texttt{ModelManager}$ component with the next grid ($G'$).  In the case of \textit{RL Brush}, the $\texttt{ModelManager}$ observes the $\texttt{ChangeEvent}$ and in turn broadcasts it to its sub-components the models.  The $\texttt{ModelManager}$ architecture will be described in more detail in a subsequent section.

\subsection{Human-Driven, AI-Augmented Design}

% \begin{figure}
%     \centering
%     \begin{subfigure}[t]{.45\linewidth}
%         \centering
%         \includegraphics[height=90px]{figures/tiles1.png}
%         \caption{The tileset for Sokoban}
%         \label{fig:tileset}
%     \end{subfigure}
%     \begin{subfigure}[t]{.45\linewidth}
%         \centering
%         \includegraphics[height=90px]{figures/tile_tools.png}
%         \caption{The edit tools.}
%         \label{fig:edit_tools}
%     \end{subfigure}
%     \caption{The tile editor tools that allow the user to modify the .}
%     \label{fig:tile_tools}
% \end{figure}

Both the $\texttt{TileEditorView}$ and the $\texttt{SuggestionView}$ respond only to user-interactions in order to ultimately provide the human in the loop the final say on whether to accept the AI suggestions or override them through manual edits.  
The goal is to provide a \textit{best-of-both-worlds} approach to human and AI co-creation in which the controls of a conventional level-editor can be augmented by AI suggestions without replacing the functionality a user would have expected from a manual tile editor.  Instead, the human drives the entire level design process while taking on a more collaborative role with the ensemble of AI level-design agents.

\subsection{\texttt{ModelManager} Data Flow}

\begin{figure}
    \centering
    \includegraphics[width=1.0\linewidth]{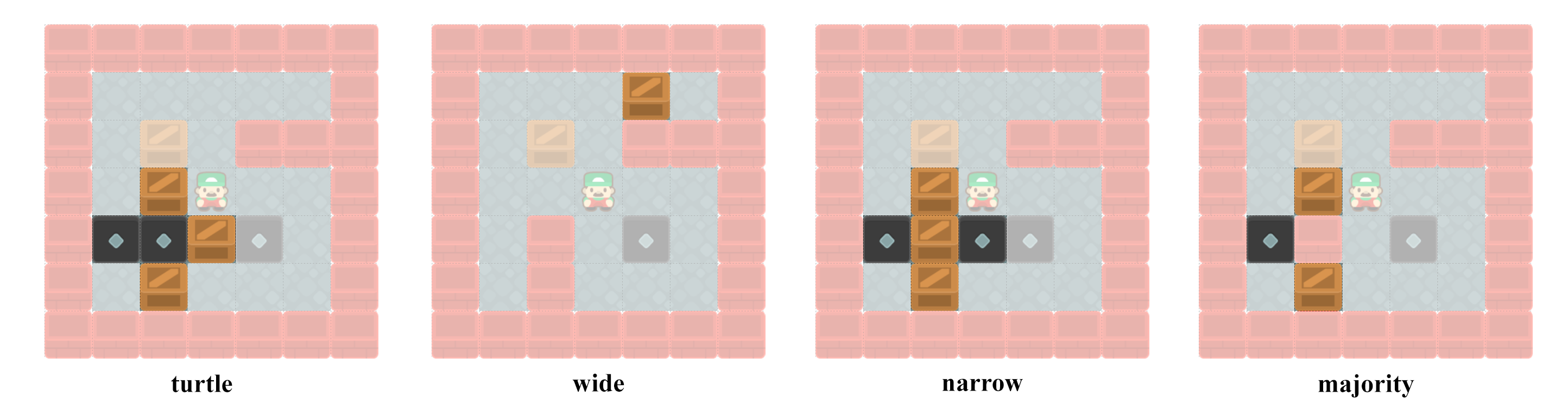}
    \caption{Each suggestion in the UI is generated by a different agent whose name appears below its diff rendering.  Clicking on the suggestion applies it to the grid $G$.}
    \label{fig:suggestions_ui}
\end{figure}

The $\texttt{ModelManager}$ in figure~\ref{fig:architecture_diagram} handles the interactions with the PCGRL agents $a = \begin{bmatrix} a_{0} & a_{1} & ... & a_{x} \end{bmatrix}$ (where $x$ is the number of used PCGRL agents) and meta-agents $m = \begin{bmatrix} m_{0} & m_{1} & ... & m_{y} \end{bmatrix}$ (where $y$ is the number of used meta-agents). The $\texttt{ModelManager}$ gets the current level state and sent to these agents where they edit it then it emits a stream of \texttt{SuggestedGrid} objects $G = \begin{matrix} G_{0} & G_{1} & ... & G_{x+y} \end{matrix}$. The $\texttt{SuggestionView}$ in turn observes the stream of $\textbf{G}$ lists and uses them to generates suggestions $\textbf{s}$ from $\textbf{G}$ by diffing them against the current level state $G_{\texttt{GridView}}$ to generate a list of suggestions $s = \begin{bmatrix} s_{0} & s_{1} & ... & s_{x+y} \end{bmatrix}$ for rendering and presenting the user in the UI's suggestion box (figure~\ref{fig:suggestions_ui}).

% TODO: Describe the PCGRL models in a bit more detail here.

% TODO: define "meta" agents.
Meta-agents in $\textbf{m}$ consist of agents that combine or aggregate the results of $\textbf{a}$ in some way to generate their results.  In \textit{RL Brush}, the \textit{majority} agent is an example of a meta-agent that aggregates one or more of the agents suggestions ($\begin{bmatrix} G_{a_{i}} & G_{a_{i+1}} & ... & G_{a_{j}} \end{bmatrix}$) to a new suggestion ($G_{m_{i}}$).  The \textit{majority} meta-agent is powered by a pure, rule-based model that only makes a suggestion of a tile mutation if the majority of the agents have the same tile mutation in their suggestions. In our case, we are using 3 different PCGRL agents (narrow, turtle, and wide) which means at least 2 agents have to agree on the same tile mutation.  

% TODO: clean up these equations to be more consistent with the diagram's notation

% The \textit{majority} meta agent $m_{\text{maj}}$ can also be thought of as a linear combination of the outputs of other agents.  More generally, $m_{\text{maj}}(t_{G}, \textbf{a})$ can be described with by equation~\ref{eq:majority_model_rule}, in which a tile \textit{t} in grid \textbf{G} is considered in its output if the agents \textbf{a} can be combined using some weighting scheme \textbf{w} and their resulting value is greater than some threshold $T$.

% \begin{equation}\label{eq:majority_model_rule}
%     m_{\text{maj}}(t_{G}, \textbf{a}) = \begin{cases}
%         1 & \text{if $\textbf{w}^{\top}\textbf{a} \geq T$} \\
%         0 & \text{if $\textbf{w}^{\top}\textbf{a} < T$}
%     \end{cases}
% \end{equation}

% More sophisticated meta agents can build on this notion and will be further explored in the discussion section.

% TODO: add transition language here.

\subsection{\texttt{ModelManager}'s Hyper-Parameters}

\begin{figure}
    \centering
    \begin{subfigure}[t]{.45\linewidth}
        \centering
        \includegraphics[height=0.25\linewidth]{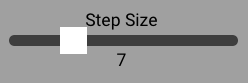}
        \caption{}
        \label{fig:stepper}
    \end{subfigure}
    \begin{subfigure}[t]{.45\linewidth}
        \centering
        \includegraphics[height=0.25\linewidth]{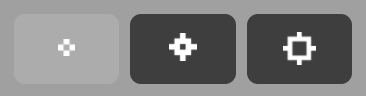}
        \caption{}
        \label{fig:tile_neighborhood}
    \end{subfigure}
    \caption{These two UI elements \textit{a} and \textit{b} control the \textit{step}  and \textit{tile radius} parameters respectively.}
    \label{fig:hyper_parameter_ui}
\end{figure}

Two primary hyper-parameters exist in \textit{RL Brush} for tuning the performance of \texttt{ModelManager}.  One is the number of \textit{steps} and the other is the \textit{tool radius}. These are each controlled from the UI using the components in figure~\ref{fig:hyper_parameter_ui}.

\begin{figure}
    \centering
    \includegraphics[width=0.3\linewidth]{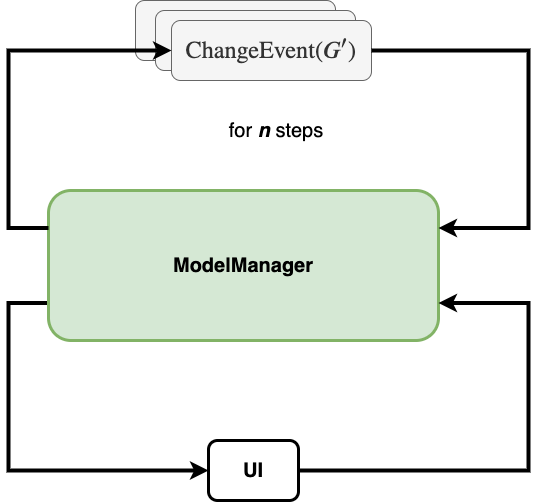}
    \caption{The changes in the \textit{step} parameter control the number of iterations \textit{n} in the loop of recursive \texttt{ChangeEvent} objects that feed back into the \texttt{ModelManager}}
    \label{fig:stepper_data_flow}
\end{figure}

The \textit{step} parameter controls how many times the \texttt{ModelManager} will call itself recursively (figure~\ref{fig:stepper_data_flow}).  For each \textit{step} the \texttt{ModelManager} will call itself recursively \textit{n} times on a self-generated stream of \texttt{ChangeEvent} (G') objects. Having a higher step value allows agents to make more than one modification to the map. This is an important hyper-parameter because most of these agents are trained to not be greedy and try to do modification that requires long term edits. Limiting these agents to only see one step ahead will suffocate them and their suggestions might not be very interesting for the users.

\begin{figure}
    \centering
    \includegraphics[width=0.7\linewidth]{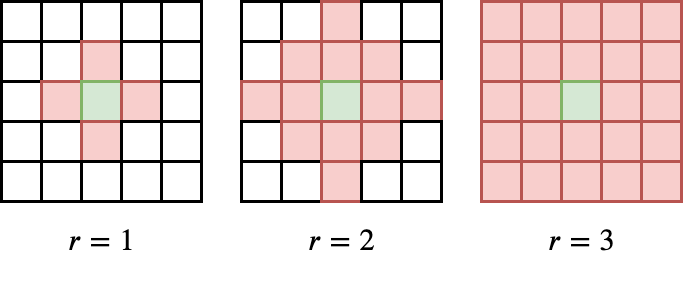}
    \caption{The changes in the \textit{tool radius} parameter control the size of the slice of grid \textit{G'} that is visible to the agents as input.}
    \label{fig:tool_radius_neighborhood}
\end{figure}

The \textit{tool radius} parameter controls how big the window of tiles are visible to the agent as input. Agents can't provide suggestions outside of this window. It focuses the suggestion to be around the area the user is modifing at the current step.  In figure~\ref{fig:tool_radius_neighborhood} the white tiles are padded as empty or as walls, depending on the agent.  The red tiles represent the integer values of each tile on the grid \textit{G}. The green tile represents the \textit{pivot tile} or position on the grid \textit{G} that the user last clicked on if a tile was added manually.  In cases where no tile was clicked\footnote{Such as the case in which the user accepted an AI suggestion}, the center of the grid \textit{G} is used as the \textit{pivot tile}.  The radius $r$ refers to the Von-Neuman neighborhood's radius with respect to the \textit{pivot tile}.  However, note that for all grids \textit{G} where $r \geq \lc\frac{\textit{gridRadius}}{2}\rc$, the entire grid is used such as in cases of $r = 3$ on \textit{microbans} of size $5 \times 5$. 

% \textit{RL Brush} demonstrates a fraction of the possibile configurations using 3 agents trained using by the PCGRL models and the 1 meta agent to generate 4 suggestions at a time for each mutation of the game board presented to the user for review (Fig. \ref{fig:suggestions_ui}) .  Humans can continuously make manual edits and are free to ignore the suggestions or they can click on a suggestion in the UI to accept it and immediately apply it to their in-progress level.  Each change to the level whether made manually through human editing or by accepting a suggestion triggers a new batch of suggestions to be generated and the entire data flow to continue in a cyclical manner until the desired level is complete.

\section{Experiments}

% TODO: add more high-level summary.  Maybe a statement of the key hypothesis?
% AHMED: I added some stuff

In this section we demonstrate through a user study conducted to study the interactions between users and the AI suggestions. We are primarily interested in answering the following four questions:
\begin{itemize}
    \item Q1: Do users prefer to use the AI suggestions or not? 
    \item Q2: Does the AI guide users to designing more playable levels?
    \item Q3: Which AI suggestions yield higher engagement from users?
    %julian please rephrase above: can't understand the sentence -- DONE, changed language to better reflect our results.
    % \item Does the AI increase user engagement with the tool?
    \item Q4: What is the effect of the AI suggestions on the playable levels?
\end{itemize}

\begin{table}
    \centering
    \begin{tabularx}{0.8\linewidth} { 
      | >{\hsize=1.5\hsize}X 
      | >{\raggedleft\arraybackslash\hsize=0.5\hsize}X | }
         \hline
          \multicolumn{2}{|c|}{ Total Event Counts} \\
          \hline
         Total User Sessions &	75 \\
         \hline
         Total Interaction Events &	3165 \\
         \hline
         Total Ghost Suggestions Accepted &	308 \\
         \hline
           \multicolumn{2}{|c|}{ Aggregations } \\
          \hline
        Level Versions Per Session &	10.6 \\
        \hline
        Ghost Suggestions Accepted Per User Session	 & 4.11 \\
        \hline
        Total Interactions Per Session &	42.2 \\
        \hline
    \end{tabularx}
    \caption{Interaction Event Summary}
    \label{fig:interaction_event_summary}
\end{table}

\begin{figure}
    \begin{subfigure}[b]{.24\linewidth}
        \centering
        \includegraphics[width=\linewidth]{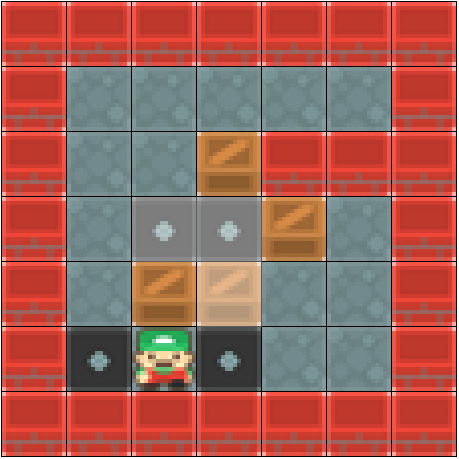}
    \end{subfigure}
        \begin{subfigure}[b]{.24\linewidth}
        \centering
        \includegraphics[width=\linewidth]{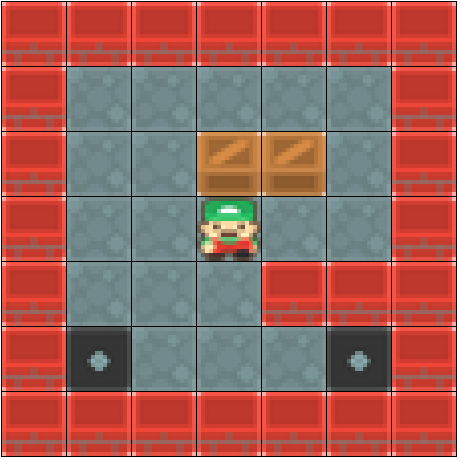}
    \end{subfigure}
        \begin{subfigure}[b]{.24\linewidth}
        \centering
        \includegraphics[width=\linewidth]{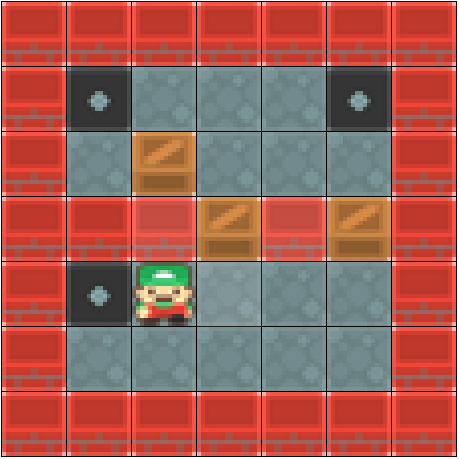}
    \end{subfigure}
    \begin{subfigure}[b]{.24\linewidth}
        \centering
        \includegraphics[width=\linewidth]{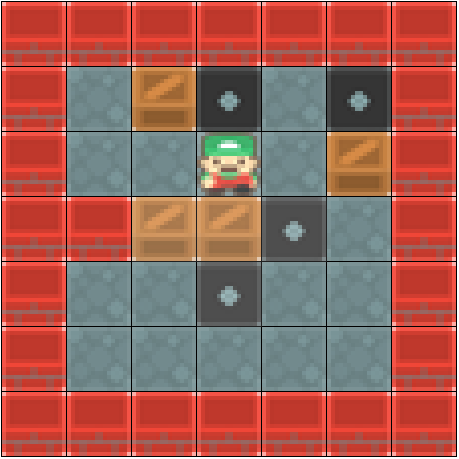}
    \end{subfigure}
    \caption{User-generated levels}
    \label{fig:user_levels}
\end{figure}

For the experiment, we published the \textit{RL Brush} app~\footnote{https://rlbrush.app/} to the web, and shared the link with university students and faculty on a shared Slack channel as well as on social media platforms Twitter and Facebook. We then recorded anonymized user-interaction events to the application web server. During the course of about 2 weeks, 75 unique user sessions were logged in total.  Figure~\ref{fig:user_levels} shows the final states of a few levels created using a combination of human and edits and AI suggestions in \textit{RL Brush}. Table \ref{fig:interaction_event_summary} shows the counts of key metrics that we used to measure the interactions of users and the \textit{RL Brush} UI.  For instance, each session resulted in an average of 10.6 level versions, defined as unique levels, throughout each user's total average of 42.2 interactions with the UI (i.e. button presses or clicks) during the course of the session.  For example, a user may have generated 2 level versions and 100 interactions in a session by making a single edit and just clicking "Undo" and "Redo" over and over again.  From these 10.6 level versions 4.11 were generated using the AI suggested edits or \textit{ghost suggestions}.

%omar: i think this section should be omitted from this draft.  if we get more results later, we can revisit this

% We also asked the users to do a questionnaire that asks them if the system was enjoyable with the AI or without it or doesn't matter. 
% %julian Was that the exact phrasing? Was that two choices or three? 
% Also, we ask them for any additional comments about the system. We ended up only having 5 total user doing the actual questionnaire with 2 people prefer the AI, 2 people prefer the system without AI, and one didn't care. We didn't get much comments about the system except for one.
% %julian I wonder if we should even publish the results from the questionnaire, given that so few people did it - maybe just say that we had a questionnaire but too few people completed it for it to be useful
% Some 
% %julian how many? this is still out of five?
% people were complaining that the AI was always suggesting to have one avatar and not allowing more than avatar on the screen. We understand that having more than one avatar can create interesting new levels but the used models where already trained to only have one player which forced the system toward erasing any extra ones.

\section{Results}

\begin{table}
    \centering
    \begin{tabular}{|p{.2\textwidth}|>{\centering\arraybackslash}p{.2\textwidth}|>{\centering\arraybackslash}p{.2\textwidth}||>{\centering\arraybackslash}p{.2\textwidth}|}
         \hline
                    & Used AI & Didn't Use AI & Total\\
         \hline
         \hline
         Playable   &   9  & 2  & 11 \\
         \hline
         UnPlayable &   8  & 20 & 28 \\
         \hline
         \hline
         Total      &   17 & 22 & 39\\
         \hline
    \end{tabular}
    \caption{Statistics on the 39 full session}
    \label{tab:session_statistics}
\end{table}

From these 75 user sessions, 39 sessions were fully logged interaction events during the session from start to finish. We analyzed these sessions on an event-by-event basis and found a few trends. Table~\ref{tab:session_statistics} shows the statistics about all these 39 fully-logged, sessions. The amount of people that didn't use the AI ($22$) is slightly higher than the ones used the AI ($17$). There might be a lot of different reasons that users never engaged with the system, but we suspect the absence of a formal tutorial could have impacted the results here. On the other hand, users that interacted with at least one AI suggestion yielded at more playable levels ($9$ out of $17$) than users did not interact with AI suggestions at all ($2$ out of $22$). There is multiple different factors that could reflect this higher percentage. We think that the main factor is that the AI suggestions nudge/inspire users toward building playable levels.

% \begin{figure}
%     \centering
%     \includegraphics[width=0.5\linewidth]{user_study/fig1_average_time_to_valid_board.png}
%     \caption{Users who did not use any AI suggestions seemed to take longer to create valid board states.}
%     \label{fig:time_to_valid_board}
% \end{figure}

% One such trend shows that, of those users who had at least one valid, solvable board during their session, the users that interacted with at least one AI suggestion create a valid board earlier in their session than those who did not use any AI suggestions, as illustrated in figure~\ref{fig:time_to_valid_board}. Perhaps the higher learning curve of using AI suggestions makes it less immediately obvious to users than the directness of manual edits. 
% Conversely, this could mean that having AI suggestions in the system makes users more engaged overall.
% Omar: This last sentence is not really proven by the data.  

% Ahmed: If possible add a graph or table that compare the average number of interactions or histograms of people who use AI vs people that doesn't use AI

\begin{figure}
    \centering
    \begin{subfigure}[t]{.45\linewidth}
        \centering
        \includegraphics[width=\linewidth]{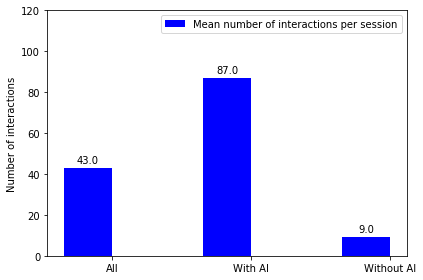}
        \caption{Mean interactions by user cohort.}
        \label{fig:mean_interactions_by_cohort}
    \end{subfigure}
    \begin{subfigure}[t]{.45\linewidth}
        \centering
        \includegraphics[width=\linewidth]{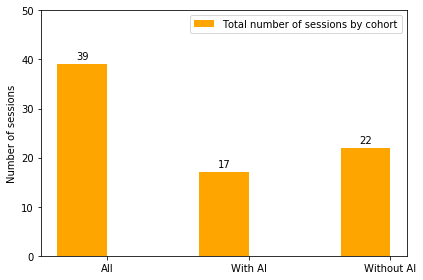}
        \caption{Total sessions by user cohort.}
        \label{fig:mean_sessions_by_cohort}
    \end{subfigure}
    \caption{The number of interactions on average is greater in the cohort of users that accepted at least one AI suggestion during their session.}
    \label{fig:session_stats}
\end{figure}

One such trend, as described in figure~\ref{fig:mean_interactions_by_cohort}, users working alongside AI (i.e. users that accepted at least one AI suggestion in their workflow) generate a higher average number of interactions when compared with the other cohort of users working without AI and the combined average of all users. The higher rate of interaction could be attributed to user-engagement, if interpreted positively, or perhaps, if interpreted negatively, that AI increases the complexity of the workflow and requires more clicks to fix any unwanted AI actions.  Since the overall average number of ghost suggestions accepted per session is 4.11, as shown on table~\ref{tab:session_statistics}, we interpret the increase in the interactions to positive engagement than an overwrought system, which we assume would have a higher number of accepted suggestions overall compared to average level versions of 10.6 as frustrated users might fight with the system and produce many more interactions per level version.  Of course, other external factors could explain the trend. The small number of users that interacted with the AI at all, as seen in figure~\ref{fig:mean_sessions_by_cohort}, could point to a majority of users new to level editing and without having a formal tutorial may have quit before discovering the AI features at all.

% One such trend shows that, of those users who had at least one valid, solvable board during their session, the users that interacted with at least one AI suggestion create a valid board earlier in their session than those who did not use any AI suggestions, as illustrated in figure~\ref{fig:time_to_valid_board}. Perhaps the higher learning curve of using AI suggestions makes it less immediately obvious to users than the directness of manual edits. 

\begin{figure}[t]
    \centering
    \includegraphics[width=0.5\linewidth]{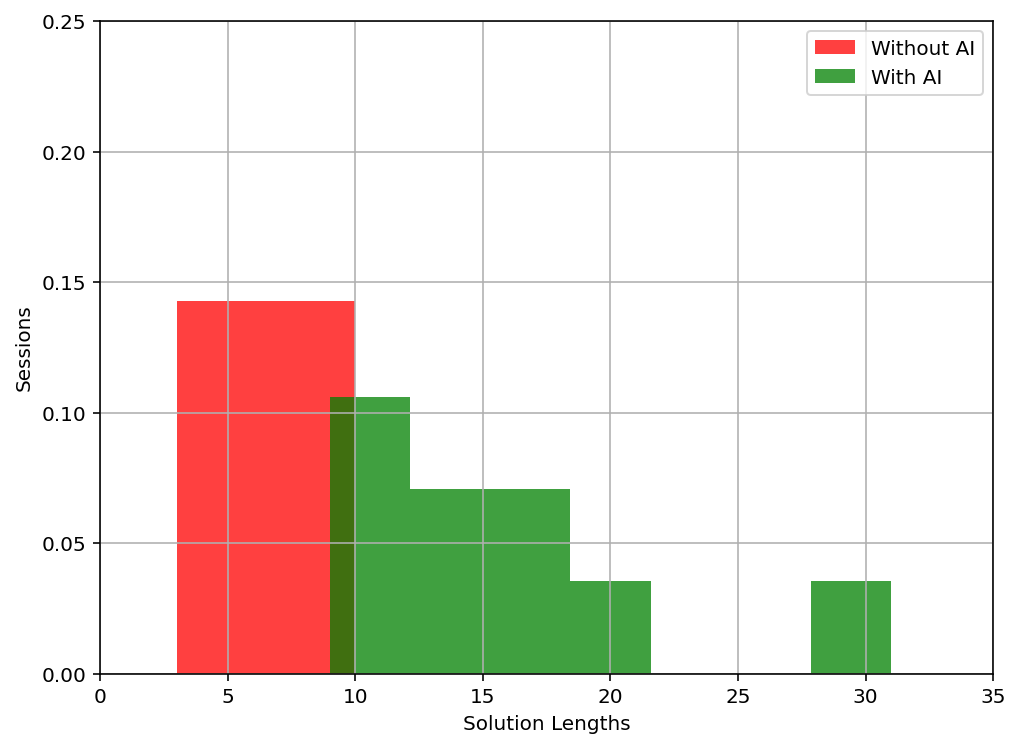}
    \caption{Users that used AI suggestions seemed to create levels that required more steps in their solutions.}
    \label{fig:level_complexity}
\end{figure}

Another trend seems to be a relationship between level complexity and and using AI suggestions, as seen in figure~\ref{fig:level_complexity}.  There the solution length is calculated using a BFS (Breadth-First Search) solver.  Each level created with the assistance of AI is, on average, longer than levels without AI. This could indicate that AI suggestions yield direct users toward creating more complex levels and therefore higher quality levels. However, further studies on a larger set of users would need to be done to further explore this trend more definitively.

\begin{figure}[t]
    \centering
    \includegraphics[width=0.5\linewidth]{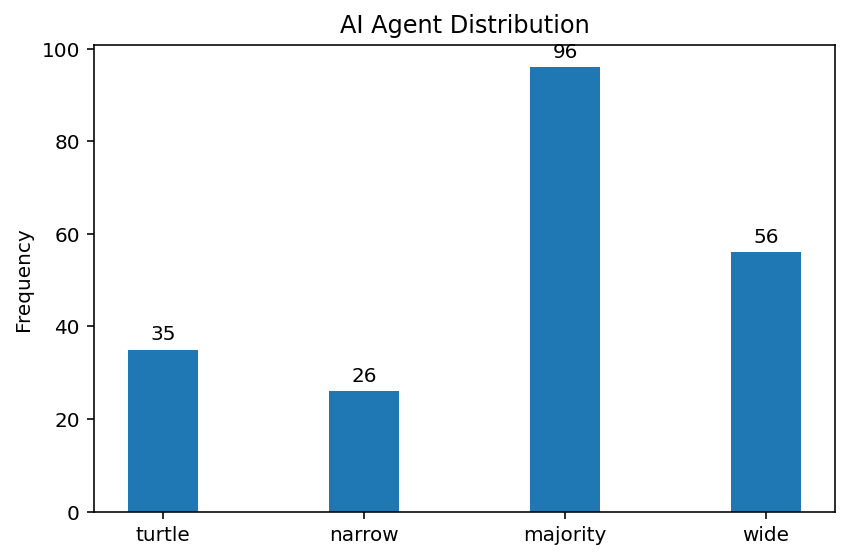}
    \caption{The \textit{majority} agent seems to be the most popular across sessions and received the most total interactions or clicks.}
    \label{fig:agent_interactions}
\end{figure}

Since \textit{RL Brush} provides different models to choose from, we were also curious to check which suggestions were most useful for the users. Figure~\ref{fig:agent_interactions} shows a histogram about which model saw the most usage, as measured by number of interactions.  We found out that the suggestions generated by the majority-voting meta-agent received the most interactions.  One interpretation for its unexpected popularity could be that agents, in isolation, may have divergent lower-quality suggestions but collectively tend toward higher suggestion quality.  In the \textit{Discussion} section, we discuss plans for further investigations aggregated suggestions.

\begin{figure}
    \centering
    \includegraphics[width=0.5\linewidth]{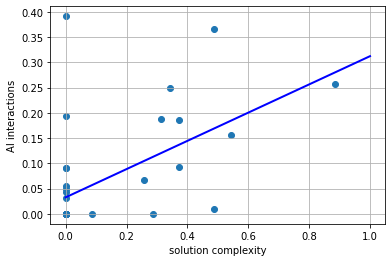}
    \caption{Correlation between number of AI suggestions accepted and overall solution complexity.}
    \label{fig:suggestions_complexity_correlation}
\end{figure}

Finally, we compute the correlation between the number of AI-accepted suggestions in a session and the solution length of the created level. Figure~\ref{fig:suggestions_complexity_correlation} shows a correlation (with coefficient equal to $0.279$) between the number of AI suggestions used during level creation and the maximum level difficulty achieved during that session, in terms of solution length. This correlation could further make the case that AI suggestions nudge users towards creating more complex levels with longer solutions.

\section{Discussion}

The system described here can be seen as a proof of concept for the idea of building mixed-initiative interaction on PCG methods based on sequential decisions. Most search-based PCG methods, as well as most PCGML methods, outputs the whole level (or other type of content) as a unit. PCGRL, where the generator has learned to design one step at a time, might afford a mode of interaction more suited to how a human designs a level. It would be interesting to investigate whether tree search approaches to level generation could be harnessed similarly~\cite{bhaumik2019tree}.

Looking back at the results, we can say that \textit{RL Brush} introduces a few areas for further exploration surrounding the relationship between user engagement and perhaps the complexity of playable levels. We also noticed that more users seem to interact more with meta-agent compared to other models. Comparing these results with our questions introduced in the Experiments section:
\begin{itemize}
    \item Q1: Based on the data we have, we can't clearly say if the users preferred to use the system with AI or without.  However, we suspect that a further study could answer this question.
    \item Q2: From the collected statistics the amount of playable levels generated by users that incorporated AI into their workflow exceeds the number of those that did not interact with the AI suggestions at all.
    \item Q3: The majority agent received the most interactions when compared to all the rest. Further studies could explore new ideas for additional types of meta-agents in future work.
    \item Q4: The results lean towards AI suggestions yielding higher quality levels, as defined using complexity of levels and with longer solution lengths, but more data would be needed to verify that.
\end{itemize}

% TODO: rewrite this in accordance to the current state of the paper.

In addition to the results described in the previous section, a broader test of human users could further explore the quality of the levels generated beyond the scope of automated solvers and through the use of human play-testing.  Additional metrics can be gathered to support this and more targeted, supervised user research can be done here. A more supervised user research will help us understand the different factor affecting our results. We would know if external factor such as game literacy and familiarity with games and level editor affects the user their engagement with AI. We believe that users with higher game literacy may find the tool less intimidating than ones with lower game literacy. Another important study could be to understand the influence of the AI suggestion on the final created levels: were the AI suggestions pivotal for the final created levels or merely inspiration for heavily hand-made levels?

Once the broader user studies have been conducted, additional client-side models can be added to \textit{RL Brush} that learn the weights of meta-agents and continuously optimize them through online-model training.  In this way, we could better leverage the \texttt{ModelManager}'s ensemble architecture's capabilities.  Furthermore, the existing PCGRL models could be extended to continuously train online using reward functions incorporating parameters based on user actions.   Similarly, novel client-side models specifically tailored to improve the UX (user experience) could be incorporated into future versions that better leverage the capabilities of TensorFlow.js, which \textit{RL Brush} utilizes in its code already.

Subsequent versions would also add support for additional games, level-design agent types and $N \times M$ grids in order to increase the overall utility of \textit{RL Brush} as a functional design tool.

% TODO: mention the generalization of meta-agents as "ensemble" models/systems.

% Additional features of \textit{RL Brush} can be used to explore broader questions about human-AI co-creation tools.  For instance, does providing the user the ability to have more overrides over the AI-assisted tool have a discernible impact on the results?  Regarding parameter tuning, how does altering the number of level updates per interaction impact the results?   Finally, do particular constraints on the brushes on where it would or wouldn't work have meaningful impact on the resulting level? 

\section{Conclusion}

In the previous sections we have introduced how \textit{RL Brush} provides a way to seamlessly integrate human level editing with AI suggestions with an opt-in paradigm.  The results of the user study suggest that using the AI suggestions in the context of level editing could impact the quality of the resulting levels.  In general, using AI suggestions seemed to result in more highly playable levels per session and higher overall level quality, as measured by solution length.

There is clearly more work to do in this general discussion. We don't know yet to what types of levels and other content this method can be applied, and there are certainly other types of interaction possible with an RL-trained incremental PCG algorithm. \textit{RL Brush} will hopefully serve as a nexus of discovery in the space of using PCGRL in game-level design.

% \section*{Acknowledgments}
% We would like to thank the reviewer agents for reading the paper, and making editing suggestions so that we can improve it in a mixed-initiative manner.
\bibliographystyle{splncs04}
\bibliography{main}

\begin{thebibliography}{10}
\providecommand{\url}[1]{\texttt{#1}}
\providecommand{\urlprefix}{URL }
\providecommand{\doi}[1]{https://doi.org/#1}

\bibitem{alvarez2018assessing}
Alvarez, A., Dahlskog, S., Font, J., Holmberg, J., Johansson, S.: Assessing
  aesthetic criteria in the evolutionary dungeon designer. In: Proceedings of
  the 13th International Conference on the Foundations of Digital Games.
  pp.~1--4 (2018)

\bibitem{alvarez2019empowering}
Alvarez, A., Dahlskog, S., Font, J., Togelius, J.: Empowering quality diversity
  in dungeon design with interactive constrained map-elites. In: 2019 IEEE
  Conference on Games (CoG). pp.~1--8. IEEE (2019)

\bibitem{bhaumik2019tree}
Bhaumik, D., Khalifa, A., Green, M.C., Togelius, J.: Tree search vs
  optimization approaches for map generation. arXiv preprint arXiv:1903.11678
  (2019)

\bibitem{butler2013mixed}
Butler, E., Smith, A.M., Liu, Y.E., Popovic, Z.: A mixed-initiative tool for
  designing level progressions in games. In: Proceedings of the 26th annual ACM
  symposium on User interface software and technology. pp. 377--386 (2013)

\bibitem{chen2018q}
Chen, Z., Amato, C., Nguyen, T.H.D., Cooper, S., Sun, Y., El-Nasr, M.S.:
  Q-deckrec: A fast deck recommendation system for collectible card games. In:
  Computational Intelligence and Games. IEEE (2018)

\bibitem{goldman2004turtle}
Goldman, R., Schaefer, S., Ju, T.: Turtle geometry in computer graphics and
  computer-aided design. Computer-Aided Design  \textbf{36}(14),  1471--1482
  (2004)

\bibitem{guzdial2019friend}
Guzdial, M., Liao, N., Chen, J., Chen, S.Y., Shah, S., Shah, V., Reno, J.,
  Smith, G., Riedl, M.O.: Friend, collaborator, student, manager: How design of
  an ai-driven game level editor affects creators. In: Proceedings of the 2019
  CHI Conference on Human Factors in Computing Systems. pp. 1--13 (2019)

\bibitem{guzdial2018co}
Guzdial, M., Liao, N., Riedl, M.: Co-creative level design via machine
  learning. arXiv preprint arXiv:1809.09420  (2018)

\bibitem{khalifa2020pcgrl}
Khalifa, A., Bontrager, P., Earle, S., Togelius, J.: Pcgrl: Procedural content
  generation via reinforcement learning. arXiv preprint arXiv:2001.09212
  (2020)

\bibitem{liapis2013sentient}
Liapis, A., Yannakakis, G.N., Togelius, J.: Sentient sketchbook:
  computer-assisted game level authoring  (2013)

\bibitem{machado2019pitako}
Machado, T., Gopstein, D., Nealen, A., Togelius, J.: Pitako-recommending game
  design elements in cicero. In: 2019 IEEE Conference on Games (CoG). pp.~1--8.
  IEEE (2019)

\bibitem{shaker2013evolving}
Shaker, N., Shaker, M., Togelius, J.: Evolving playable content for cut the
  rope through a simulation-based approach. In: Ninth Artificial Intelligence
  and Interactive Digital Entertainment Conference (2013)

\bibitem{shaker2013ropossum}
Shaker, N., Shaker, M., Togelius, J.: Ropossum: An authoring tool for
  designing, optimizing and solving cut the rope levels. In: Ninth Artificial
  Intelligence and Interactive Digital Entertainment Conference (2013)

\bibitem{shaker2016procedural}
Shaker, N., Togelius, J., Nelson, M.J.: Procedural content generation in games.
  Springer (2016)

\bibitem{smith2010tanagra}
Smith, G., Whitehead, J., Mateas, M.: Tanagra: A mixed-initiative level design
  tool pp. 209--216 (2010)

\bibitem{summerville2018procedural}
Summerville, A., Snodgrass, S., Guzdial, M., Holmg{\aa}rd, C., Hoover, A.K.,
  Isaksen, A., Nealen, A., Togelius, J.: Procedural content generation via
  machine learning (pcgml). IEEE Transactions on Games  \textbf{10}(3),
  257--270 (2018)

\bibitem{summerville2016mystical}
Summerville, A.J., Mateas, M.: Mystical tutor: A magic: The gathering design
  assistant via denoising sequence-to-sequence learning. In: Twelfth artificial
  intelligence and interactive digital entertainment conference (2016)

\bibitem{sutton1998introduction}
Sutton, R.S., Barto, A.G., et~al.: Introduction to reinforcement learning,
  vol.~135. MIT press Cambridge (1998)

\bibitem{torrado2019bootstrapping}
Torrado, R.R., Khalifa, A., Green, M.C., Justesen, N., Risi, S., Togelius, J.:
  Bootstrapping conditional gans for video game level generation  (2019)

\bibitem{yannakakis2014mixed}
Yannakakis, G.N., Liapis, A., Alexopoulos, C.: Mixed-initiative co-creativity
  (2014)

\bibitem{yannakakis2018artificial}
Yannakakis, G.N., Togelius, J.: Artificial intelligence and games, vol.~2.
  Springer (2018)

\bibitem{zhu2018explainable}
Zhu, J., Liapis, A., Risi, S., Bidarra, R., Youngblood, G.M.: Explainable ai
  for designers: A human-centered perspective on mixed-initiative co-creation.
  In: 2018 IEEE Conference on Computational Intelligence and Games (CIG).
  pp.~1--8. IEEE (2018)

\end{thebibliography}





% LNCS Homepage, \url{http://www.springer.com/lncs}. Last accessed 4
% \end{thebibliography}
\end{document}